# Neural Modelling and Control of a Diesel Engine with Pollution Constraints

**Mustapha Ouladsine\*, Gérard Bloch\*\*, Xavier Dovifaaz\*\***

\* LSIS, Domaine Universitaire de Saint-Jérôme (UMR CNRS 6168)
Avenue de l'Escadrille Normandie Niemen, 13397 Marseille Cedex 20, France
Email: mustapha.ouladsine@univ.u-3mrs.fr
\*\* Centre de Recherche en Automatique de Nancy (CRAN, UMR CNRS 7039)
CRAN-ESSTIN, 2 rue Jean Lamour, 54500 Vandoeuvre, France
Email: gerard.bloch@esstin.uhp-nancy.fr

**Abstract:** The paper describes a neural approach for modelling and control of a turbocharged Diesel engine. A neural model, whose structure is mainly based on some physical equations describing the engine behaviour, is built for the rotation speed and the exhaust gas opacity. The model is composed of three interconnected neural sub-models, each of them constituting a nonlinear Multi-Input Single-Output Output Error model. The structural identification and the parameter estimation from data gathered on a real engine are described. The neural direct model is then used to determine a neural controller of the engine, in a specialized training scheme minimising a multivariable criterion. Simulations show the effect of the pollution constraint weighting on a trajectory tracking of the engine speed. Neural networks, which are flexible and parsimonious nonlinear black-box models, with universal approximation capabilities, can describe or control accurately complex nonlinear systems, with few a priori theoretical knowledge. The presented work extends optimal neuro-control to the multivariable case and shows the flexibility of neural optimisers. Considering the preliminary results, it appears that neural networks can be used as embedded models for engine control, to satisfy the more and more restricting pollutant emission legislation. Particularly, they are able to model nonlinear dynamics and outperform during transients the control schemes based on static mappings.

## 1. Introduction

Neural techniques are used in various domains and particularly for system modelling and control (Chen, Billings and Grant, 1990; Narendra and Parthasarathy, 1990; Pham, 1995; Nørgaard *et al.*, 2000). Neural networks bring important benefits by suppressing theoretical difficulties that appear when applying classical techniques on complex systems. Including nonlinearities in their structure, they can describe or control accurately complex nonlinear systems, with few a priori theoretical knowledge. In (Bloch and Denoeux, 2003), the advantages of neural models are summarized: they are *flexible and parsimonious nonlinear black-box models*, with *universal approximation capabilities*.

In this paper, neural techniques are applied to model and control a turbocharged Diesel engine. The objective is to build a model to be used to control the Diesel engine. The engine speed and the exhaust gas opacity that characterizes one type of pollution must be controlled. More precisely, the control should allow reducing the opacity peaks that occur during engine acceleration. Neural networks are used because they can approximate and replace the complex and nonlinear thermodynamical, mechanical and chemical equations that describe the Diesel engine (Cook *et al.*, 1996). For Diesel engines control, neural networks have been already used (Hafner *et al.*, 1999). Other works use neural predictors to optimise Air-Fuel Ratio (AFR) control (Majors *et al.*, 1994; Magner and Jankovic, 2002; Bloch *et al.*, 2003). The work presented here is application oriented, just as the papers cited above. It extends optimal neuro-control to the multivariable case and shows the flexibility of neural optimisers. Considering the preliminary results, it appears that neural networks can be used as embedded models for engine control, to satisfy the more and more restricting pollutant emission legislation. Particularly, they are able to model nonlinear dynamics and outperform during transients the control schemes based on static mappings.

Section 2 presents the building of the neural model for the rotation speed and the exhaust gas opacity. The structure of this model is mainly based on some physical equations describing the engine behaviour. The final model is composed of three interconnected neural sub-models, each of them constituting a nonlinear Multi-Input Single-Output (MISO) Output Error (OE) model. The structural identification (i.e. the determination of the regressor structure and the internal architecture) and the parameter estimation from data gathered on a real engine are described. Experimental results are then presented.

In section 3, the (direct) model obtained in the previous section is used to determine a neural controller of the engine, in a specialized training scheme based on the minimisation of a multivariable criterion. The simulation of a trajectory tracking of the engine speed with and without pollution constraints is finally presented.

## 2. Neural modelling

### 2.1. A neural model of a Diesel engine

A turbocharged Diesel engine can be decomposed into subsystems as presented in Figure 1. The atmospheric air goes through the turbocharger compressor, the air intake manifold, and the combustion chamber. The injection pump injects fuel in the combustion chamber while the valves are closed, and the mixture burns. The gases produced by the explosion pass through the exhaust manifold and turbocharger turbine and are ejected out away. Five states have been modelled: the engine speed $R$, the intake manifold pressure $P$, the inlet airflow $\dot{m}$, the fuel flow $\dot{m}_f$ and the opacity $Op$ of the exhaust gas. This work is mainly focused on the engine speed and exhaust gas opacity. The only control variable considered here is the position $T$ of the injection pump.

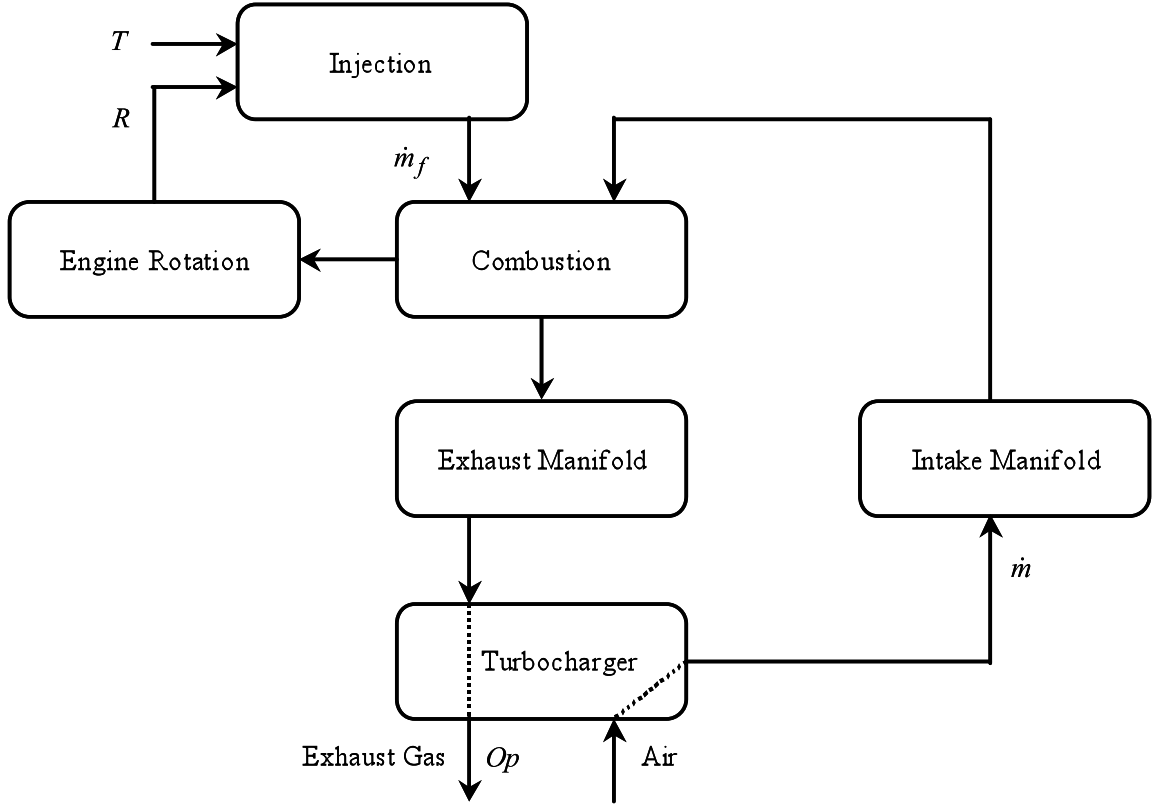

Figure 1: Turbocharged Diesel engine.

The physical relations that describe the behaviour of an internal combustion engine are used to design the corresponding neural model. Among the different possibilities, the Diesel engine speed $R$, the intake manifold pressure $P$ and the exhaust gas opacity $Op$ can be described with the following formal relations:

$$\frac{dR(t)}{dt} = f_R(R(t), P(t), T(t))$$
$$\frac{dP(t)}{dt} = f_P(R(t), P(t)) \qquad (1)$$
$$Op(t) = f_{Op}(R(t), \dot{m}(t), \dot{m}_f(t))$$

It is worth noting that the speed dynamics is mainly due to the engine inertia and that the speed value naturally depends on the injected fuel rate and thus on the injection pump position $T$. On the other hand, the opacity depends on the ratio between the air and fuel masses (AFR) used in the combustion, and then on the flows of air $\dot{m}$ and fuel $\dot{m}_f$.

These relations are used to construct the neural model used to estimate the speed, the pressure and the opacity. The first step consists in discretizing the previous equations, which gives:

$$R(k) = F_R(R(k-1),\cdots,R(k-n_{RR}),P(k-1),\cdots,P(k-n_{PR}),T(k-1),\cdots,T(k-n_{TR}))$$
$$P(k) = F_P(R(k-1),\cdots,R(k-n_{RP}),P(k-1),\cdots,P(k-n_{PP})) \qquad (2)$$
$$O_P(k) = F_{Op}(R(k),\dot{m}(k),\dot{m}_f(k))$$

where $F_R$, $F_P$ and $F_{Op}$ are models in discrete time $k$ of the engine speed, the intake manifold pressure and the exhaust opacity, respectively, and where $n_{RR}$, $n_{PR}$, $n_{TR}$, $n_{RP}$ and $n_{PP}$ are orders that must be identified.

Some modifications of the opacity model are needed. First, the opacity is measured at the exhaust of the Diesel engine. This means that there is some pure delay between the opacity measurement and the other variables and that there are some dynamics due to the gas transportation. Secondly, the opacity depends on the injected fuel flow $\dot{m}_f$. Some works (Blanke and Andersen, 1985) show that this quantity mainly depends on the engine speed $R$ and on the injection pump position $T$:

$$\dot{m}_f(k) = f(R(k),T(k)) \qquad (3)$$

The opacity can be then expressed as:

$$Op(k) = F_{Op}(Op(k-1),\cdots,Op(k-n_{Op}),T(k-d),R(k-d),\dot{m}(k-d)) \qquad (4)$$

where $d$ is the delay mentioned above and $n_{Op}$ the order of the opacity model.

One objective of the modelling is to control the engine. However, the complete model obtained is not easy to use for control because of the interdependency of the speed and pressure models. Some simplifications have been thus carried out. On one hand, the speed at time $k$ depends on the control variable $T$, the speed $R$ and the pressure $P$, at previous times. On the other hand, this pressure $P$ depends on speed $R$ at previous times. It is then possible to express the speed as a function depending on the control $T$ and speed $R$ only. Finally, the engine speed, the pressure and the opacity are given by:

$$R(k) = F_R(R(k-1),\cdots,R(k-n_{RR}),T(k-1),\cdots,T(k-n_{TR}))$$
$$P(k) = F_P(R(k-1),\cdots,R(k-n_{RP}),P(k-1),\cdots,P(k-n_{PP})) \qquad (5)$$
$$O_P(k) = F_{Op}(Op(k-1),\cdots,Op(k-n_{Op}),R(k-d),T(k-d),\dot{m}(k))$$

The three unknown models $F_R$, $F_P$ and $F_{Op}$ were estimated from data by using neural structures, more precisely conventional MLP (Multi-Layer Perceptron) architectures with one hidden layer of sigmoidal units and linear activation function for the output unit (see (Bloch and Denoeux, 2003), for example). The training (i.e. estimation of the parameters) was performed by minimizing the mean squared error function, with a batch Levenberg-Marquardt algorithm (see (Nørgaard *et al.*, 2000), for example). Note that each neural model is identified in the form of a Multi-Input Single-Output (MISO) Output Error (or simulation) model, involving, among the regressors, delayed estimates, not delayed measurements, of the output.

To complete the models, the orders of the regressors (i.e. the lag space) and the number of nodes in the hidden layer must be determined for each sub-model. The neural structure selection used is presented on Figure 2. For each order value, the neural network is trained with a given node number, chosen large enough, and the Final Prediction Error (FPE) criterion is calculated. The optimal order corresponds to the minimal FPE value. The neural network is then trained with this order, but for several values of the node number in the hidden layer. Again an analysis of the FPE criterion gives the optimal node number corresponding to the minimal FPE value and thus the final network structure.

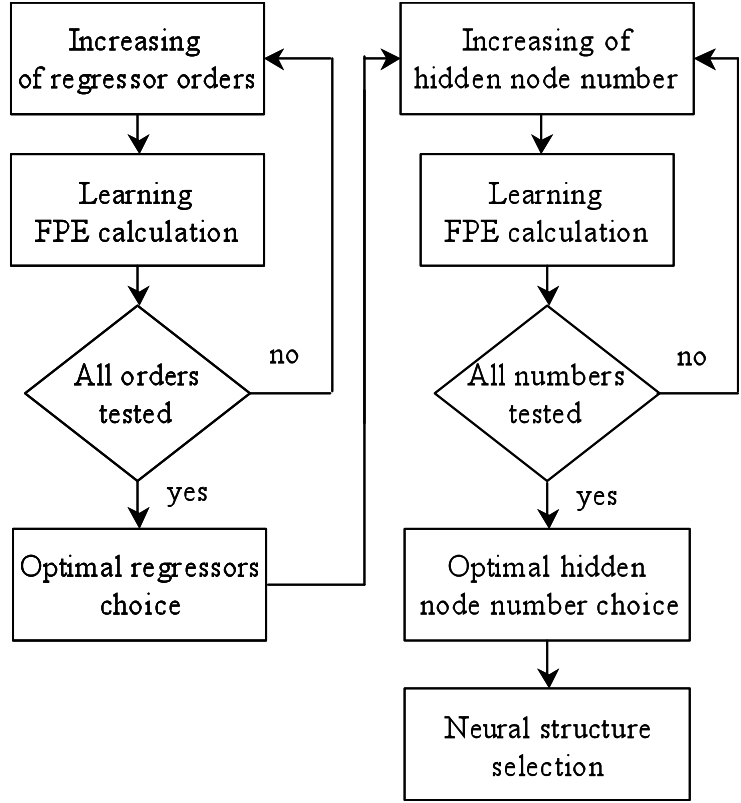

Figure 2: Structural identification process.

This process is repeated for each network, leading to the final models of speed and opacity:

$$\hat{R}(k) = NN_R\left(\hat{R}(k-1), \hat{R}(k-2), T(k-1)\right)$$
$$\hat{P}(k) = NN_P\left(\hat{P}(k-1), \hat{R}(k-1)\right) \quad (6)$$
$$O_P(k) = NN_{OP}\left(O_P(k-1), T(k-4), \hat{R}(k-4), \dot{m}(k-4)\right)$$

The complete model consists then in three interconnected neural networks. Each neural network is a multilayer perceptron composed of several inputs, an output and a single hidden layer. One of them reconstructs the engine speed *R* from the control *T*. The external input of the pressure model is the speed model output. The opacity is estimated from the speed and airflow estimates and the control variable. These recurrent neural networks have been trained using data of the control variable *T*, the speed *R*, the airflow $\dot{m}$ and the opacity *Op*. The complete simulation model is then obtained by connecting the networks.

## 2.2. Experimental results

The following figures present some results obtained from a real Diesel engine. Figures 3, 4 and 5 present the measurements and estimations respectively for the speed, pressure and opacity, for data used to train the neural models (identification data). In order to validate the model, another time sequence for the input *T* was applied to the real system and the neural model. The corresponding measurements and estimations of speed, pressure and opacity are given in Figures 6, 7 and 8.

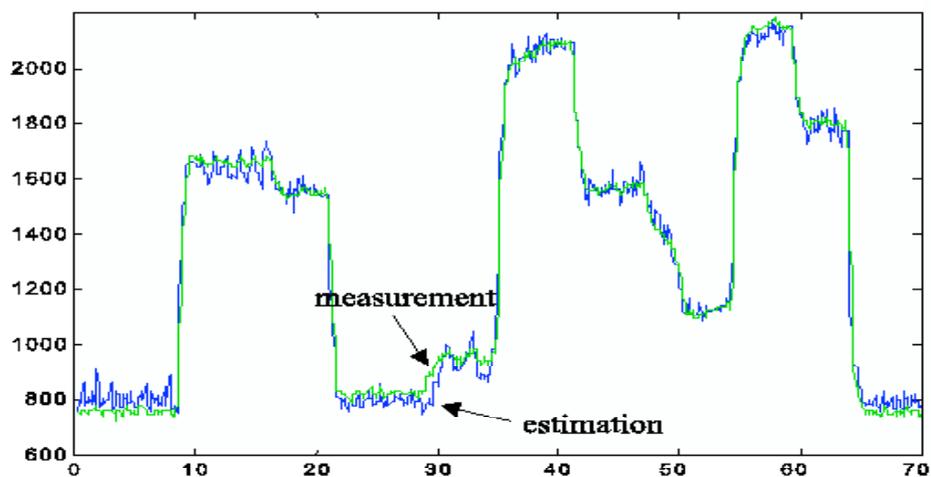

Figure 3: Measurements and estimates of speed (rpm), identification data.

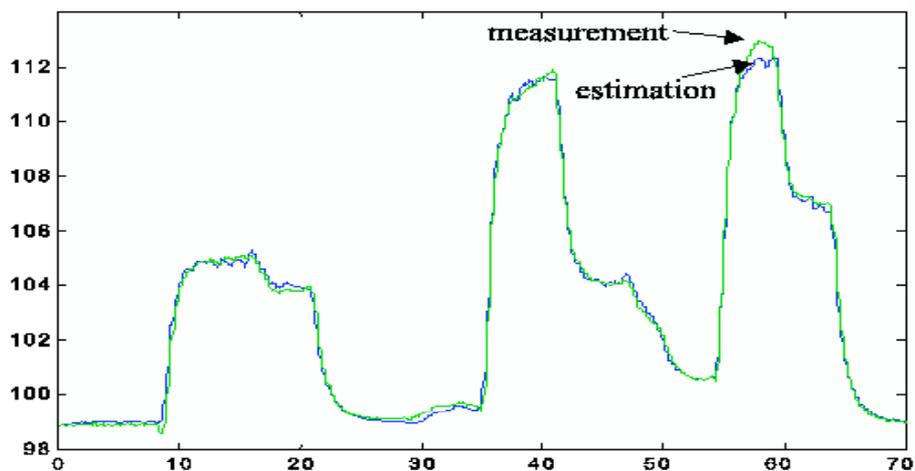

Figure 4: Measurements and estimates of pressure (kPa), identification data.

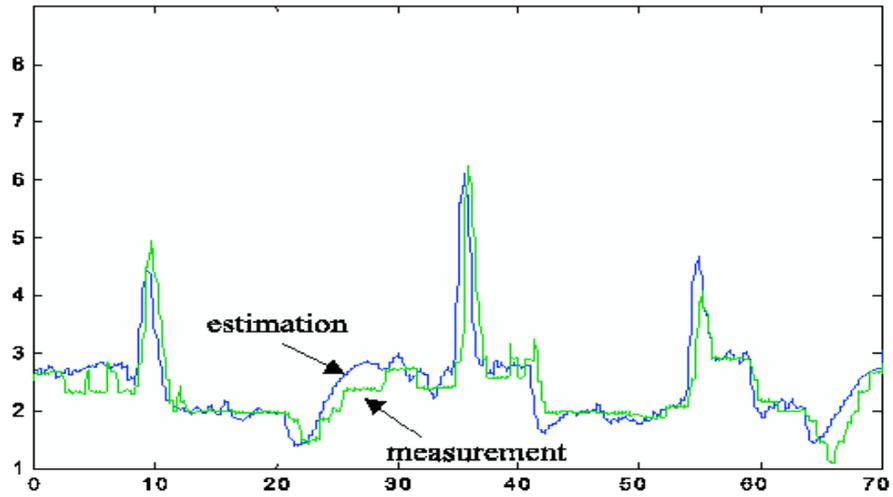

Figure 5: Measurements and estimates of opacity (%), identification data.

Even if the estimations are given by a complete simulation neural model whose single input is the injection pump position $T$, the estimates of speed, pressure and opacity are very close to the measurements. The engine model reproduces the static and the dynamical behaviour of the system with a good precision. Figures 5 and 8 show that peaks and static levels of opacity are well estimated, despite the dynamics and nonlinearities.

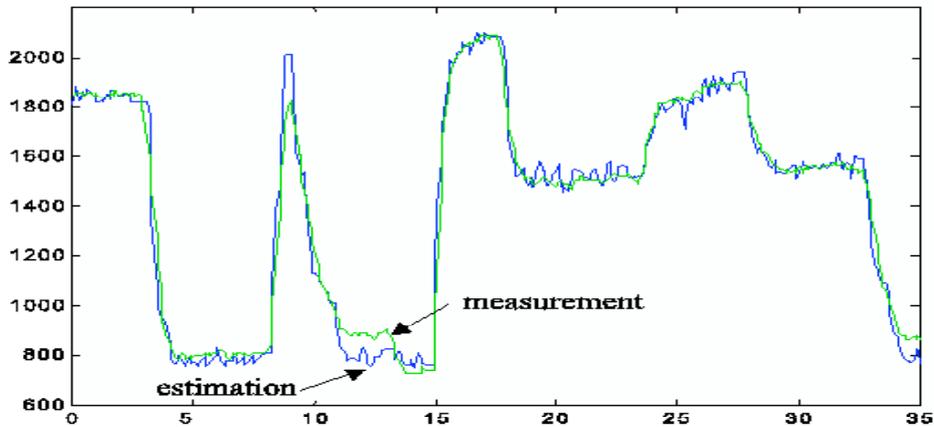

Figure 6: Measurements and estimates of speed, validation data.

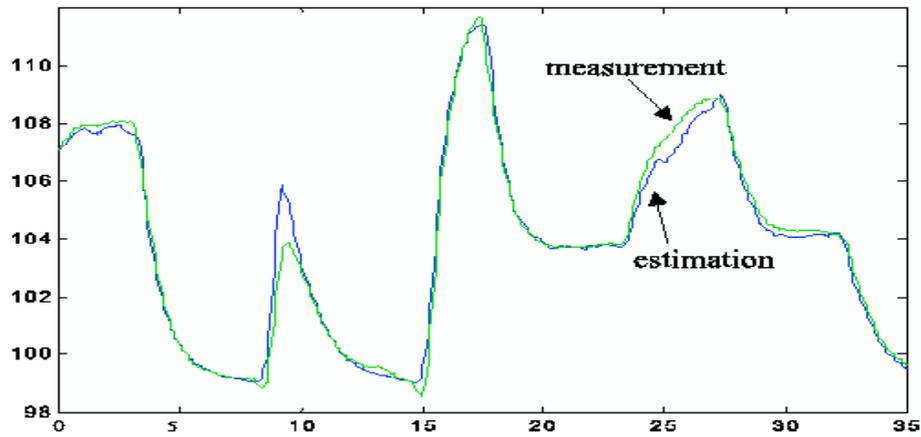
Figure 7: Measurements and estimates of pressure (kPa), validation data.

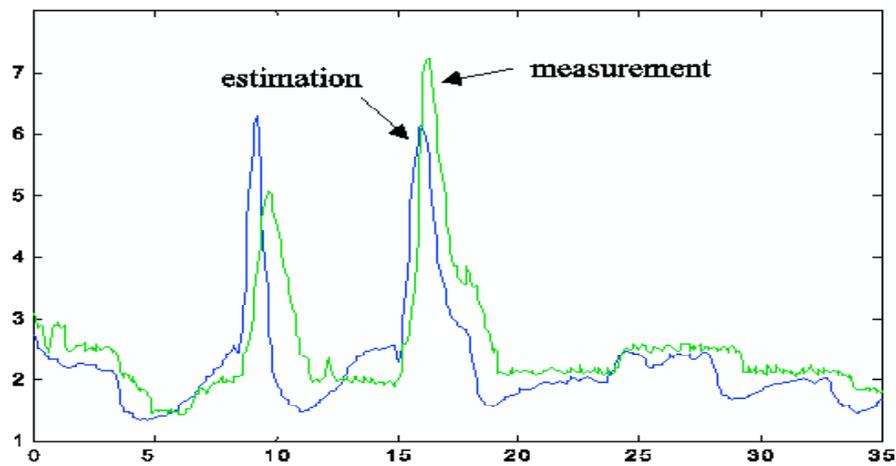
Figure 8: Measurements and estimates of opacity (%), validation data.

## 3. Neural control

### 3.1. Introduction and theory

There is presently a vast literature on neuro-control and successive states of the art have been regularly provided (Hunt *et al.*, 1992; Narendra, 1996; Narendra and Lewis, 2001). As almost all linear control schemes can be extended to nonlinear systems by using neural networks, various neuro-control schemes can be found, like inverse control (He *et al.*, 1999), internal model control (Hunt and Sbarbaro, 1991; Rivals and Personnaz, 2000), predictive control (Eaton *et al.*, 1994; Soloway and Haley, 1996; Chen *et al.*, 1999; Vila and Wagner, 2003), optimal control (Plumer, 1996) and adaptive control (Chen and Khalil, 1995). The control can be direct or indirect, depending on whether a neural model of the system is needed or not. In the presented application, the stability issue of the control scheme is not considered. The approach used for constructing the control is indirect since a neural network model of the engine is used. The controller is a neural network that has to be trained to satisfy the objective of speed tracking with opacity constraint. The training scheme used is the specialized training, generally credited to Psaltis (Psaltis *et al.*, 1988), detailed in (Nørgaard *et al.*, 2000). Its principle is displayed on Figure 9.

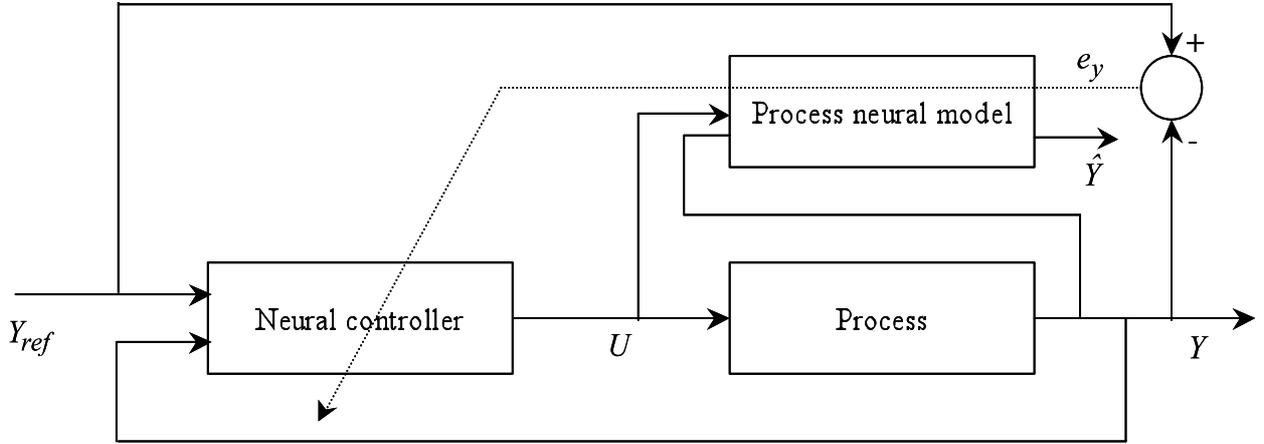

Figure 9: Architecture of training for control.

The control parameters (neural controller weights) are generally learned to minimize the sum of squared errors between the reference and the system output. The main difficulty lies in the fact that the minimization algorithm requires the Jacobian of the system, i.e. the derivatives of the output with respect to the input, which, most of the time, is unknown. This problem is overcome by including a neural (direct) model of the system in the training scheme and estimating the Jacobian from this model.

This method was applied to control the engine with, as direct model, the neural model presented in the previous section. The criterion, adapted to include the engine speed and the opacity, is then a multivariable criterion. However, for sake of simplicity, the algorithm, applied in the multivariable case for engine control, is first detailed for a criterion containing only one variable.

To tune the neural controller parameters, a recursive algorithm is used, in an approach very close to adaptive control. The criterion is defined at discrete time *t* by:

$$J_t(W) = \frac{1}{2} \sum_{k=1}^{t} e_y(W,k)^2 \qquad (7)$$

where $e_y(W,k) = Y_{ref}(k) - Y(W,k)$, $Y_{ref}(k)$ and $Y(k)$ are respectively the desired output, i.e. the reference, and the actual output of the system, at time *k*.

The general rule used to update the weights is as follows:

$$W^t = W^{t-1} - \mu_t \, [R_t(W^{t-1})]^{-1} J'_t(W^{t-1}) \qquad (8)$$

where $W^t$ denotes the controller parameter vector $W = (w_1 \; w_2 \cdots w_n)^T$ updated at time *t*, $\mu_t$ the step length. The gradient $J'(W)$ of the criterion is defined by:

$$J'(W) = \left( \frac{\partial J(W)}{\partial w_1} \quad \frac{\partial J(W)}{\partial w_2} \quad \ldots \quad \frac{\partial J(W)}{\partial w_n} \right)^T \tag{9}$$

The matrix $R(W)$ can be chosen in various ways. In the steepest descent algorithm, $R(W) = I$ (identity matrix). This algorithm is the simplest, but it is rarely efficient and not compatible with the differences of scales that can exist between the parameters. In the Gauss-Newton algorithm, $\mu_t = 1$ and $R(W)$ is an approximation of the Hessian (or second derivatives) matrix. Some theoretical developments can be found in (Ljung and Söderström, 1983; Nørgaard *et al.*, 2000).

The algorithm description needs to develop the first and the second derivatives of the criterion with respect to the weights. From (7):

$$\frac{\partial J_t(W)}{\partial w_i} = \frac{\partial J_{t-1}(W)}{\partial w_i} + \frac{1}{2} \frac{\partial e_y(W,t)^2}{\partial w_i} = \frac{\partial J_{t-1}(W)}{\partial w_i} + e_y(W,t) \frac{\partial e_y(W,t)}{\partial w_i} \tag{10}$$

and then:

$$J'_t(W) = J'_{t-1}(W) + e_y(W,t)\Psi_y(W,t) \tag{11}$$

where

$$\Psi_y(W,t) = \left( \frac{\partial e_y(W,t)}{\partial w_1} \quad \ldots \quad \frac{\partial e_y(W,t)}{\partial w_n} \right)^T \tag{12}$$

When applying the rule (8), the term $J'_{t-1}(W^{t-1})$ is considered to be zero since $W^t$ minimize $J_t$ at every step (Ljung and Söderström, 1983). Thus:

$$J'_t(W^{t-1}) = e_y(W^{t-1},t)\Psi_y(W^{t-1},t) \tag{13}$$

The second derivative is developed as follows:

$$\frac{\partial^2 J_t(W)}{\partial w_i \partial w_j} = \frac{\partial^2 J_{t-1}(W)}{\partial w_j \partial w_i} + \left( \frac{\partial e_y(W,t)}{\partial w_j} \frac{\partial e_y(W,t)}{\partial w_i} + e_y(W,t) \frac{\partial^2 e_y(W,t)}{\partial w_j \partial w_i} \right) \tag{14}$$

Assuming that the error $e_y$ can be considered as a white noise, not correlated with the second derivatives $\frac{\partial^2 e_y}{\partial w_j \partial w_i}$, the approximate Hessian of the criterion, at iteration $t$, $R_t$, can be written in a matrix form:

$$R_t(W) = R_{t-1}(W) + \Psi_y(W,t)\Psi_y^T(W,t) \tag{15}$$

Finally, we have:

$$\begin{cases} W^t = W^{t-1} - [R_t]^{-1} e_y(W^{t-1},t)\Psi_y(W^{t-1},t) \\ R_t = R_{t-1} + \Psi_y(W^{t-1},t)\Psi_y^T(W^{t-1},t) \end{cases} \tag{16}$$

To replace the matrix inversion with a scalar one, the classical matrix inversion lemma is used:

$$\begin{cases} W^t = W^{t-1} - P_t\, e_y(W^{t-1},t)\, \Psi_y(W^{t-1},t) \\ P_t = P_{t-1} - \dfrac{P_{t-1}\, \Psi_y(W^{t-1},t)\, \Psi_y^T(W^{t-1},t)\, P_{t-1}}{1 + \Psi_y^T(W^{t-1},t)\, P_{t-1}\, \Psi_y(W^{t-1},t)} \end{cases} \quad (17)$$

The vector $\Psi_y$, defined in (12), contains the derivatives $\dfrac{\partial e_y(W,t)}{\partial w_i}$ that are developed as follows:

$$\frac{\partial e_y(W,t)}{\partial w_i} = -\frac{\partial Y(W,t)}{\partial w_i} = -\frac{\partial Y(W,t)}{\partial U(W,t-1)}\frac{\partial U(W,t-1)}{\partial w_i} \quad (18)$$

where $U$ is the input vector of the system, i.e. the neural controller output. Replacing the Jacobian of the system by the Jacobian of the neural model gives:

$$\frac{\partial e_y(W,t)}{\partial w_i} = -\frac{\partial \hat{Y}(W,t)}{\partial U(W,t-1)}\frac{dU(W,t-1)}{dw_i} \quad (19)$$

where $\hat{Y}(W,t)$ is the model output and where $\dfrac{dU(W,t-1)}{dw_i}$ only depends on the controller structure.

The controller parameter learning method previously described is now applied to a multivariable criterion, involving two system outputs $Y$ and $Z$. Defining $e_y(W,k) = Y_{ref}(k) - Y(W,k)$ and $e_z(W,k) = Z_{ref}(k) - Z(W,k)$, the criterion can be written:

$$J(W) = \frac{1}{2}\sum_{k=1}^{t}\left(\eta_y\, e_y(W,k)^2 + \eta_z\, e_z(W,k)^2\right) \quad (20)$$

where $\eta_y$ and $\eta_y$ are weighting factors. In this case, using the same notation as before and using the matrix inversion lemma twice, the final minimisation algorithm is given by:

$$\begin{cases} W^t = W^{t-1} - P_t\left(\eta_y\, e_y(W^{t-1},t)\, \Psi_y(W^{t-1},t) + \eta_z\, e_z(W^{t-1},t)\, \Psi_z(W^{t-1},t)\right) \\ M = P_{t-1} - \dfrac{P_{t-1}\, \Psi_y(W^{t-1},t)\, \Psi_y^T(W^{t-1},t)\, P_{t-1}}{1 + \Psi_y^T(W^{t-1},t)\, P_{t-1}\, \Psi_y(W^{t-1},t)} \\ P_t = M - \dfrac{M\, \Psi_z(W^{t-1},t)\, \Psi_z^T(W^{t-1},t)\, M}{1 + \Psi_z^T(W^{t-1},t)\, M\, \Psi_z(W^{t-1},t)} \end{cases} \quad (21)$$

where $\Psi_y(W,t) = \left(\dfrac{\partial e_y(W,t)}{\partial w_1} \cdots \dfrac{\partial e_y(W,t)}{\partial w_n}\right)^T$ and $\Psi_z(W,t) = \left(\dfrac{\partial e_z(W,t)}{\partial w_1} \cdots \dfrac{\partial e_z(W,t)}{\partial w_n}\right)^T$.

### 3.2. Application to the Diesel engine

This section describes the simulation of Diesel engine control, i.e. the control application where the actual engine is replaced by its model obtained in section 2, as presented on Figure 10.

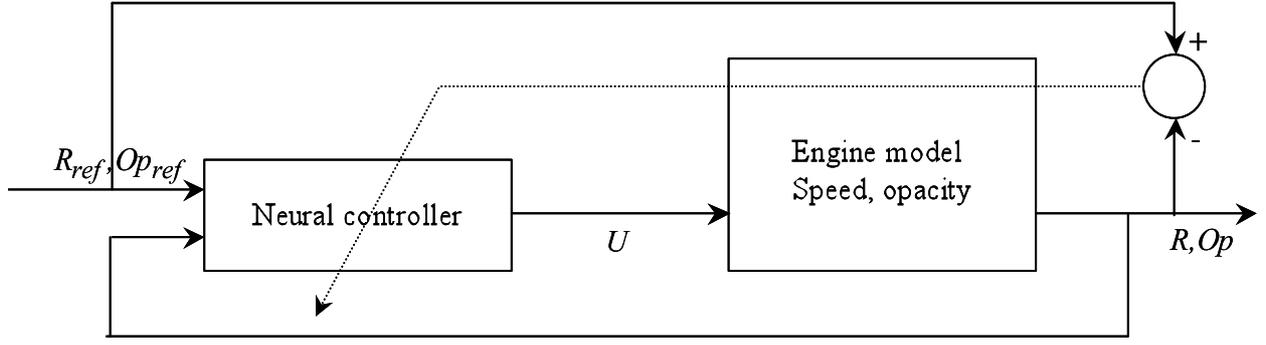

Figure 10: Training applied to the engine model.

Since the objective is to control the engine speed and the opacity, the criterion is defined by:

$$J(W) = \frac{1}{2} \sum_{k=1}^{N} \left( \left( R_{ref}(k) - R(W,k) \right)^2 + \eta_{Op} \left( Op_{ref}(k+d-1) - Op(W,k+d-1) \right)^2 \right) \quad (22)$$

where $R_{ref}$ is the speed reference and where $Op_{ref}$ represents the opacity constraint, defined such that the opacity is reduced during the transients. $\eta_{Op}$ is the weighting factor of the opacity constraint. For instance, with $\eta_{Op} = 0$, the control is a simple engine speed tracking without opacity constraint. The controller output $U$ is calculated by a MLP with one hidden layer of sigmoidal units, from inputs which are the speed and opacity references and the speed and opacity system outputs. The controller output is then expressed by a neural function of the following form:

$$U(k+1) = NN_U \left( R_{ref}(k+1), R(k), R(k-1), Op_{ref}(k+d), Op(k+d-1) \right) \quad (23)$$

Training was carried out using the algorithm presented in (21). Figures 11 and 12 show the simulation results obtained with the resulting neural controller for three values of the weighting factor, $\eta_{Op} = 0, 0.2, 0.8$, for the speed and the opacity respectively.

It is worth noting that, when $\eta_{Op} = 0$, the neural controller performs a good tracking of the engine speed, since the speed output follows precisely the reference. On the contrary, when the opacity constraint is taken into account ($\eta_{Op} \neq 0$), a tracking error occurs during the transients (acceleration). The peaks of opacity are caused by an excess of injected fuel (depending on the injection pump position $U$) during engine acceleration. Naturally, to satisfy the opacity constraint, the neural controller calculates $U$ such that less fuel is injected. This leads to a decrease of the acceleration and then to a speed tracking error during the transients. This error increases with the weighting factor $\eta_{Op}$ while in the same time the peaks of opacity decrease.

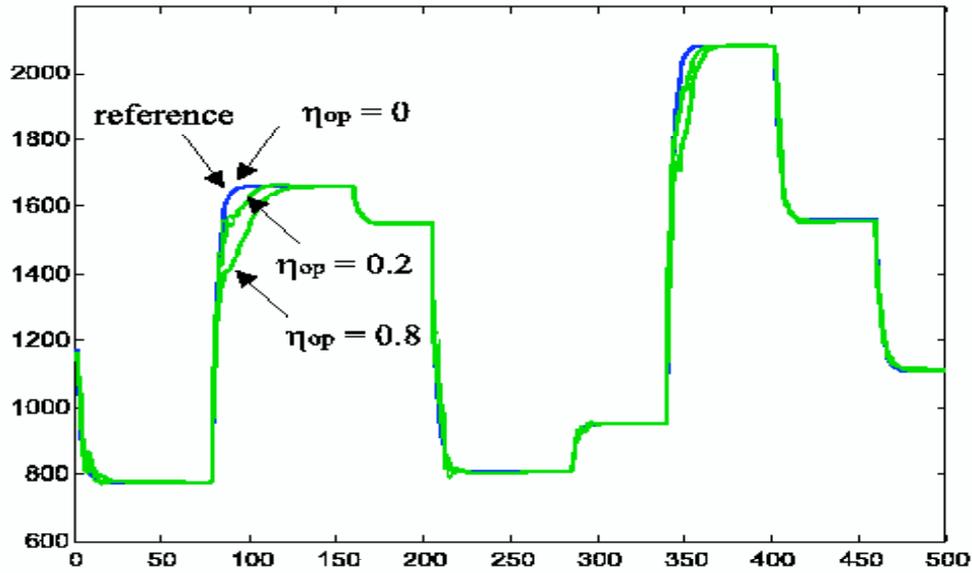

Figure 11: Engine speed (rpm) with respect to time (s).

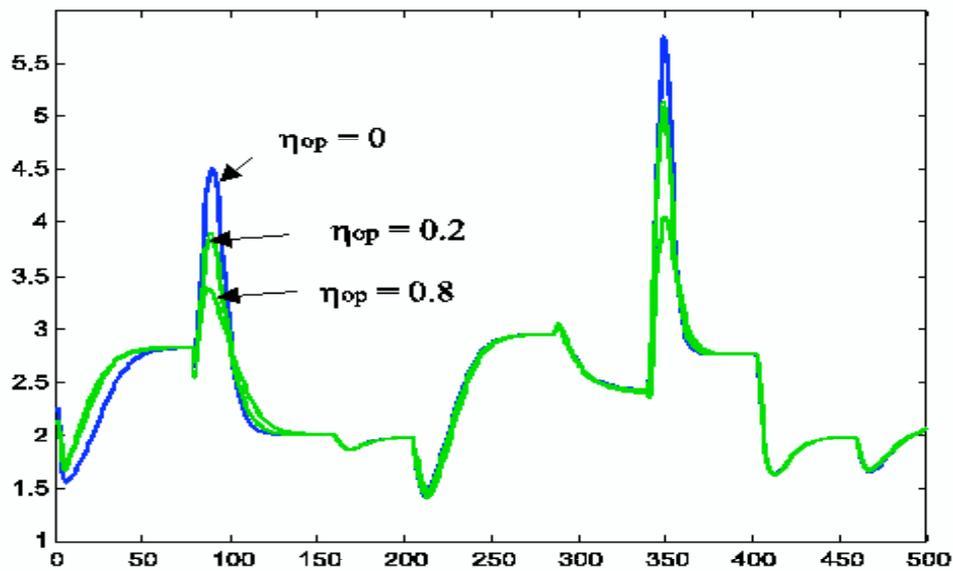

Figure 12: Opacity (%) with respect to time (s).

## 4. Conclusion

In this paper, we presented a control scheme of Diesel engine speed with pollution constraints. This scheme used the specialized training of a neural controller using a neural direct engine model. To include pollution constraints, the criterion to minimise includes both the engine speed and the exhaust gas opacity. The work extends optimal neuro-control to the multivariable case and shows the flexibility of neural optimisers. The results highlight the interest of using neural networks both for engine modelling and control, despite strong dynamics and nonlinearities (opacity). Obviously, an important work must be done to implement neural controllers on real engines. However, considering the preliminary results, it seems that neural networks can be used as embedded models for engine control, to satisfy the more and more restricting pollutant emission legislation. Particularly, they are able to model the nonlinear dynamics and outperform during transients the classical control schemes. They could constitute an interesting alternative to the methods employed for pollution control which are based on the use of set point cartographies.